\documentclass[runningheads,a4paper]{llncs}

\usepackage[colorlinks=true,citecolor=blue]{hyperref}
\usepackage{amsmath,amssymb}
\usepackage{graphicx}
\graphicspath{ {images/} }

\usepackage{multirow}
\usepackage{url}
\usepackage{verbatim}

\urldef{\mailsa}\path|rcha9612@uni.sydney.edu.au, aditya.menon@data61.csiro.au, schawla@qf.org.qa|

\newcommand{\keywords}[1]{\par\addvspace\baselineskip
\noindent\keywordname\enspace\ignorespaces#1}

\usepackage{subfigure}
\usepackage{multirow}
\usepackage{soul,color,colortbl}
\usepackage{xcolor}
\usepackage{booktabs}
\usepackage{upgreek}

\newcommand{\E}{\mathbf{E}}
\newcommand{\I}{\mathbf{I}}

\newcommand{\N}{\mathbf{N}}

\renewcommand{\S}{\mathbf{S}}

\newcommand{\U}{\mathbf{U}}
\newcommand{\V}{\mathbf{V}}
\newcommand{\W}{\mathbf{W}}
\newcommand{\X}{\mathbf{X}}
\newcommand{\Z}{\mathbf{Z}}

\newcommand{\Real}{\mathbb{R}}

\let\oldFootnote\footnote
\newcommand\nextToken\relax

\renewcommand\footnote[1]{%
    \oldFootnote{#1}\futurelet\nextToken\isFootnote}

\newcommand\isFootnote{%
    \ifx\footnote\nextToken\textsuperscript{,}\fi}

\begin{document}

\mainmatter  

\title{Robust, Deep and Inductive Anomaly Detection}

\titlerunning{Robust, Deep and Inductive Anomaly Detection}

%
%
\author{Raghavendra Chalapathy\inst{1} \and Aditya Krishna Menon\inst{2} \and Sanjay Chawla\inst{3}}

\authorrunning{Chalapathy, Menon and Chawla}

\institute{University of Sydney and Capital Markets Cooperative Research Centre (CMCRC)
\and 
Data61/CSIRO and the Australian National University
\and
Qatar Computing Research Institute (QCRI), HBKU\\
\mailsa\\
 }

%
%

\toctitle{Lecture Notes in Computer Science}
\tocauthor{Authors' Instructions}
\maketitle

\begin{abstract}

PCA is a classical statistical technique whose simplicity and maturity has seen it find widespread use for anomaly detection.
However, it is limited in this regard by being sensitive to gross perturbations of the input, and by seeking a linear subspace that captures normal behaviour.
The first issue has been dealt with by \emph{robust PCA}, a variant of PCA that explicitly allows for some data points to be arbitrarily corrupted;
however, this does not resolve the second issue,
and indeed introduces the new issue that one can no longer inductively find anomalies on a test set.
This paper addresses both issues in a single model, the \emph{robust autoencoder}.
This method learns a nonlinear subspace that captures the majority of data points, while allowing for some data to have arbitrary corruption.
The model is simple to train and leverages recent advances in the optimisation of deep neural networks.
Experiments on a range of real-world datasets highlight the model's effectiveness.


\end{abstract}

\section{Anomaly detection: motivation and challenges}

A common need when analysing real-world datasets is determining which instances stand out as being dramatically dissimilar to all others.
Such instances are known as \emph{anomalies}, and the goal of \emph{anomaly detection} (also known as \emph{outlier detection}) is to determine all such instances in a data-driven fashion~\cite{chandola2007outlier}.
Anomalies can be caused by errors in the data but sometimes are indicative of a new, previously unknown, underlying process;
in fact Hawkins~\cite{hawkins} defines an outlier as an observation that {\it deviates so significantly from other observations as to arouse suspicion that it was generated by a different mechanism.}

Principal Component Analysis (PCA) \cite{Hotelling:1933} is a core method for a range of statistical inference tasks, including anomaly detection.
The basic idea of PCA is that while many data sets are high-dimensional, they tend to inhabit a {low-dimensional manifold}.
PCA thus operates by (linearly) projecting data into a lower-dimensional space, so as to separate the {\em signal} from the {\em noise};
a data point which is far away from its projection is deemed as anomalous.

While intuitive and popular, PCA has limitations as an anomaly detection method.
Notably, it is highly sensitive to data perturbation: one extreme data point can completely change the orientation of the projection, often leading to the masking of anomalies.
A variant of PCA, known as a \emph{robust} PCA (RPCA) limits the impact of anomalies by using a clever decomposition of the data matrix~\cite{candes2010robust}.
We will discuss RPCA in detail in Section~\ref{sec:background},
but note here that it still carries out a linear projection,
and further cannot be used to make predictions on test instances;
that is, we cannot perform \emph{inductive} anomaly detection.

In this paper, we will relax the linear projection limitation of RPCA by using a deep and robust autoencoder~\cite{vincent2010stacked,Goodfellow-et-al-2016}.
The difference between RPCA and a deep autoencoder will be the use of a nonlinear activation function and the potential use of several hidden layers in the autoencoder. 
While this modification is conceptually simple, we show it yields noticeable improvements in anomaly detection performance on complex real-world image data, where a linear projection cannot capture sufficient structure in the data.
Further, the robust autoencoder is capable of performing inductive anomaly detection, unlike RPCA.

In the sequel,
we provide an overview of anomaly detection methods (Section~\ref{sec:related}), with a specific emphasis on matrix decomposition techniques
such as PCA and its robust extensions.
We then proceed to describe our proposed model based on autoencoders (Section~\ref{sec:method}),
and present our experiment setup and results (Section~\ref{sec:experiment-setup}, \ref{sec:experiment-results}).
Finally, we describe directions for future work (Section~\ref{sec:conclusion}).

\section{Background and related work on anomaly detection}
\label{sec:background}
\label{sec:related}

Consider a feature matrix $\X \in \Real^{N \times D}$,
where $N$ denotes the number of data points and $D$ the number of features for each point.
For example, $N$ could be the number of images in some photo collection, and $D$ the number of pixels used to represent each image.
The goal of anomaly detection is to determine which rows of $\X$ are anomalous, in the sense of being dissimilar to all other rows. 
We will use $\X_{i :}$ to denote the $i$th row of $\X$.

\subsection{A tour of anomaly detection methods}

Anomaly detection is a widely researched topic in the data mining and machine learning community~\cite{chandola2007outlier,charubook}.
The two primary strands of research have been the design of novel algorithms to detect anomalies,
and the design \emph{efficient} means of discovering all anomalies in a large dataset.
In the latter strand, starting from the work of Bay and Schwabacher~\cite{bay03}, several optimisations have been proposed to discover anomalies in near linear time~\cite{Ghoting:2008}.

In the former strand, which is our primary focus, most emphasis has been on non-parametric methods like distance and density based outliers~\cite{knorr1997unified,breunig2000lof}. 
For example, distance-based methods define a domain-dependent dissimilarity metric, and deem a point to be anomalous if it is relatively far away from its neighbours~\cite{Zhao:2009}. 
Another popular approach is the one-class SVM, which learns a smooth boundary that captures the majority of probability mass of the data~\cite{Scholkopf:2001}.

In recent years,
matrix factorization methods for anomaly detection have become popular. 
These methods provide a \emph{reconstruction matrix} $\hat{\X} \in \Real^{N \times D}$ of the input $\X$, and use the norm $\| \X_{i :} - \hat{\X}_{i :} \|_2^2$ as a measure of how anomalous a particular point $\X_{i :}$ is;
if the reconstruction is close to the input, then it is deemed normal;
else, anomalous.
We describe several popular examples of this approach, beginning with principal component analysis (PCA).

\subsection{PCA for anomaly detection}

PCA finds the directions of maximal variance of the data.
Supposing without loss of generality that the data matrix $\X$ has zero mean,
this may be understood as the result of a matrix factorisation \cite{Bishop:2006}:
\begin{equation}
	\label{eqn:pca}
	\min_{\W^T \W = \I, \Z} \| \X - \W \Z \|_F^2 = \min_{\U} \| \X - \X \U \U^T \|_F^2.
\end{equation}
Here,
the reconstruction matrix is $\hat{\X} = \X \U \U^T$,
where
$\U \in \Real^{D \times K}$ for some number of \emph{latent dimensions} $K \ll D$.
We can interpret $\X \U$ as a projection (or encoding) of $\X$ into a $K$-dimensional subspace,
with the application of $\U^T$ as an inverse projection (or decoding) back into the original $D$ dimensional space.

\subsection{Autoencoders for anomaly detection}

PCA assumes a linear subspace explains the data.
To relax this assumption, consider instead
\begin{equation}
	\label{eqn:autoencoder}
	\min_{\U, \V} \| \X - f( \X \U ) \V \|_F^2
\end{equation}
for some non-decreasing \emph{activation function} $f \colon \Real \to \Real$,
and $\U \in \Real^{D \times K}, \V \in \Real^{K \times D}$.
For the purposes of anomaly detection, one can define the reconstruction matrix as $\hat{\X} = f( \X \U ) \V$.

Equation \ref{eqn:autoencoder} corresponds to an autoencoder with a single hidden layer \cite{Goodfellow-et-al-2016}.
Popular choices of $f( \cdot )$ include the sigmoid $f( a ) = (1 + \exp(-a))^{-1}$ and the rectified linear unit or ReLU $f( x ) = \max(0, a)$.
As before, we can interpret $\X \U$ as an encoding of $\X$ into a $K$-dimensional subspace; however,
by applying a nonlinear $f( \cdot )$,
the projection is implicitly onto a nonlinear manifold.

\subsection{Robust PCA}

Another way to generalise PCA is to solve, for a tuning parameter $\lambda > 0$,
\begin{equation}
	\label{eqn:robust-pca}
	\min_{\S, \N} \| \S \|_* + \lambda \cdot \| \N \|_1 : \X = \S + \N,
\end{equation}
where $\| \cdot \|_*$ denotes the trace or nuclear norm $\| \X \|_* = \mathrm{tr}( (\X^T \X)^{1/2} )$,
and $\| \cdot \|_1$ the elementwise $\ell_1$ norm.
For the purposes of anomaly detection, one can define the reconstruction matrix $\hat{\X} = \X - \N = \S$.

Intuitively, Equation \ref{eqn:robust-pca} separates $\X$ into a signal matrix $\S$ and a noise matrix $\N$,
where the signal matrix has low-rank structure, and the noise is assumed to not overwhelm the signal for most of the matrix entries.
The trace norm may be seen as a convex relaxation of the rank function;
thus, this objective can be understood as a relaxed version of PCA.

Equation \ref{eqn:robust-pca} corresponds to robust PCA (RPCA)~\cite{candes2010robust}.
Unlike standard PCA, this objective can effortlessly deal with a single entry perturbed arbitrarily.
When $\lambda \to +\infty$, we will end up with $\N = \mathbf{0}, \S = \X$,
i.e.\, we will claim that there is no noise in the data, and so all points are deemed normal.
On the other hand, when $\lambda \to 0$, we will end up with $\N = \X, \S = \mathbf{0}$,
i.e.\, we will claim that there is no signal in the data, and so points with high norm are deemed anomalous.

\subsection{Direct robust matrix factorization}

Building upon RPCA,
Xiong et. al.~\cite{xiong2011direct} introduced the direct robust matrix factorization method (DRMF),
where for tuning parameters $K, e$ one solves:
\begin{equation}
	\label{eqn:drmf}
	\begin{aligned}
	\min_{\S, \N} \hspace{0.5cm} & \| \X - (\N + \S) \|_{F}^2 \colon \mbox{rank}(\S) \leq K, \|\N\|_{0} \leq e.
	\end{aligned}	
\end{equation}
As before, the matrix $\N$ captures the anomalies and $\S$ captures the signal.
Unlike RPCA, one explicitly constraints $\S$ to be low-rank, rather than merely having low trace norm;
and one explicitly constraints $\N$ to have a maximal number of nonzeros, rather than merely having bounded $\ell_1$ norm.
The lack of convexity of the objective requires a bespoke algorithm for the optimisation.

\subsection{Robust kernel PCA}

Another way to overcome the linear assumption of PCA is the robust kernel PCA (RKPCA) approach of~\cite{Nguyen:2009}.
For a feature mapping $\Upphi$ into a reproducing kernel Hilbert space, and projection operator $\mathbf{P}$ of a point into the KPCA subspace, it is proposed to reconstruct an input $\mathbf{x} \in \Real^D$ by solving the pre-image problem
\begin{equation}
	\label{eqn:rkpca}
	\hat{\mathbf{x}} = \underset{\mathbf{z} \in \Real^D}{\mathrm{argmin}} \, E_0( \mathbf{x}, \mathbf{z} ) + C \cdot \| \Upphi( \mathbf{z} ) - \mathbf{P} \Upphi( \mathbf{z} ) \|^2,
\end{equation}
where $E_0$ is a robust measure of reconstruction error (i.e.\ not merely the Euclidean norm),
and $C > 0$ is a tuning parameter.
RKPCA does not explicitly handle gross outliers, unlike RPCA;
however, by choosing a rich feature mapping $\Upphi$, 
one can capture nonlinear anomalies.
This choice of feature mapping must be pre-specified, whereas autoencoder methods implicitly \emph{learn} a good mapping.

\section{From robust PCA to robust autoencoders}
\label{sec:method}

We now present our robust (convolutional) autoencoder model for anomaly detection.
The method can be seen as an extension of robust PCA to allow for a nonlinear manifold that explains most of the data.

\subsection{Robust (convolutional) autoencoders}

Let $f \colon \Real \to \Real$ be some non-decreasing {activation function}.
Now consider the following objective, which combines the salient elements of Equations \ref{eqn:autoencoder} and \ref{eqn:robust-pca}:
\begin{equation}
	\label{eqn:robust-ae}
	\min_{\U, \V, \N} \| \X - (f(\X \U) \V + \N) \|_F^2 + \frac{\mu}{2} \cdot (\| \U \|_F^2 + \| \V \|_F^2) + \lambda \cdot \| \N \|_1,
\end{equation}
where $f( \cdot )$ is understood to act elementwise, and $\lambda, \mu > 0$ are tuning parameters.
This is a form of \emph{robust autoencoder}:
one encodes the input into the latent representation $\Z = f( \X \U )$,
which is then decoded via $\V$.
The additional $\N$ term captures gross outliers in the data, as with robust PCA.
For the purposes of anomaly detection, we have reconstruction matrix $\hat{\X} = f(\X \U) \V$.

When $\lambda \to +\infty$, we get $\N = \mathbf{0}$, and the model reduces to a standard autoencoder (Equation \ref{eqn:autoencoder}).
When $\lambda \to 0$, then one possible solution is $\N = \X$ and $\U = \V = \mathbf{0}$, so that the model memorises the training data.
For intermediate $\lambda$, the model augments a standard autoencoder with a noise absorption term that endows robustness.

More generally, Equation \ref{eqn:robust-ae} can be seen as an instance of
\begin{equation}
	\label{eqn:robust-cae}
	\min_{\theta, \N} \| \X - (\hat{\X}( \theta ) + \N) \|_F^2 + \frac{\mu}{2} \cdot \mathrm{\Omega}( \theta ) + \lambda \cdot \| \N \|_1,
\end{equation}
where $\hat{\X}( \theta )$ is some generic predictor with parameters $\theta$, and $\mathrm{\Omega}( \cdot )$ a regularisation function.
Observe that we could pick $\hat{\X}( \theta )$ to be a convolutional autoencoder~\cite{Jain:2008,vincent2010stacked}, which would be suitable when dealing with image data;
such a model will be studied extensively in our experiments.
Further, the regulariser $\mathrm{\Omega}$ could involve more general matrix norms, such as the $\ell_{1,2}$ norm \cite{Huang:2010}.

\subsection{Training the model}
\label{sec:training}

The objective function of the model of Equation \ref{eqn:robust-ae}, \ref{eqn:robust-cae} is non-convex, but unconstrained and sub-differentiable.
There are several ways of performing optimisation.
For example, for differentiable activation $f$, one could compute sub-gradients with respect to all model parameters and apply backpropagation.
However, to leverage existing advances in training deep networks, we observe that:
\begin{itemize}
	\item For fixed $\N$, the objective is equivalent to that of a standard (convolutional) autoencoder on the matrix $\X - \N$.
	Thus, one can optimise the parameters $\theta$ using any modern (stochastic) optimisation tool for deep learning that exploits gradients, such as Adam \cite{kingma2014adam}.

	\item For fixed $\theta$ (i.e.\, $\U, \V$ in the standard autoencoder case), the objective is
	$$ \min_{\theta, \N} \| \N - (\X - \hat{\X}( \theta )) \|_F^2 + \lambda \cdot \| \N \|_1, $$
	which trivially solvable via the soft thresholding operator on the matrix $\X - \hat{\X}( \theta )$ \cite{Bach:2011}, with solution
	$$ \N_{ij} =
	\begin{cases}
		(\X - \hat{\X}( \theta ))_{ij} - \frac{\lambda}{2} & \text{ if } (\X - \hat{\X}( \theta ))_{ij} > \frac{\lambda}{2} \\
		(\X - \hat{\X}( \theta ))_{ij} + \frac{\lambda}{2} & \text{ if } (\X - \hat{\X}( \theta ))_{ij} < -\frac{\lambda}{2} \\
		0 & \text{ else. }
	\end{cases}
	$$
\end{itemize}
We thus alternately optimise $\N$ and $\theta$ until the change in the overall objective is below some threshold.
The use of stochastic optimisation for the first step, and the simplicity of the optimisation for the second step, means that we can easily train the model where data arrives in an online or streaming fashion.

\subsection{Predicting with the model}

One convenient property of our model is that the anomaly detector will be inductive, i.e.\ it can generalise to unseen data points.
One can interpret the model as learning a robust representation of the input, which is unaffected by gross noise;
such a representation should thus be able to accurately model any unseen points that lie on the same manifold as the data used to train the model.

Formally, given a new $\mathbf{x}_* \in \Real^D$, one simply computes $f( \mathbf{x}_*^T \U ) \V$ to score this point.
The larger $\| \mathbf{x}_* - \V^T f( \U^T \mathbf{x}_* ) \|_2^2$ is, the more likely the point is deemed to be anomalous. 
We emphasise that such inductive predictions are simply not possible with the robust PCA method, as it estimates parameters for the $N \times D$ observations present in $\X$, with no means of generalising to unseen data.

\subsection{Connection to robust PCA}

While the robust autoencoder of Equation \ref{eqn:robust-ae} has clear conceptual similarity to robust PCA,
it may seem that choices such as the $\ell_2$ penalty on $\U, \V$ are somewhat arbitrarily used in place of the trace norm.
We now show how the objective can in fact be naturally derived as an extension of RPCA.

The trace norm can be represented in the variational form~\cite{recht2010guaranteed}
$ \| \S \|_* = \min_{\W \V = \S} \frac{1}{2} \cdot (\| \W \|_F^2 + \| \V \|_F^2). $
The robust PCA objective is thus equivalently
$$ \min_{\W, \V, \N} \frac{1}{2} \cdot (\| \W \|_F^2 + \| \V \|_F^2) + \lambda \cdot \| \N \|_1 : \X = \W \V + \N. $$
This objective has the disadvantage of being non-convex,
but the advantage of being amenable to extensions.
Pick some $\mu > 0$, and consider a relaxed version of the robust PCA objective:
$$ \min_{\W, \V, \N, \E} \| \E \|_F^2 + \frac{\mu}{2} \cdot (\| \W \|_F^2 + \| \V \|_F^2) + \lambda \cdot \| \N \|_1 : \X = \W \V + \N + \E. $$
Here, we allow for further systematic errors $\E$
which have low average magnitude.
We can equally consider the unconstrained objective
\begin{equation}
	\label{eqn:rpca-v2}
	\min_{\W, \V, \N} \| \X - (\W \V + \N) \|_F^2 + \frac{\mu}{2} \cdot (\| \W \|_F^2 + \| \V \|_F^2) + \lambda \cdot \| \N \|_1
\end{equation}
This re-expression of robust PCA has been previously noted, for example in Sprechmann et al.~\cite{Sprechmann:2015}.
To derive the robust autoencoder from Equation \ref{eqn:rpca-v2}, suppose now that we constrain $\W = \X \U$.
This is a natural constraint in light of Equation \ref{eqn:pca}, since for standard PCA we factorise $\X$ into $\hat{\X} = \X \U \U^T$.
Then, we have the objective
$$ \min_{\U, \V, \N} \| \X - (\X \U \V + \N) \|_F^2 + \frac{\mu}{2} \cdot (\| \X \U \|_F^2 + \| \V \|_F^2) + \lambda \cdot \| \N \|_1. $$
Now suppose we modify the regulariser to only operate on $\U$ rather than $\X \U$:
$$ \min_{\U, \V, \N} \| \X - (\X \U \V + \N) \|_F^2 + \frac{\mu}{2} \cdot (\| \U \|_F^2 + \| \V \|_F^2) + \lambda \cdot \| \N \|_1. $$
This is again natural in the context of standard PCA, since there we have $\W = \X \U$ satisfying $\W^T \W = \I$.
Observe now that we have derived Equation \ref{eqn:robust-ae} for a linear activation function $f( x ) = x$.
The robust autoencoder thus extends this model by employing a nonlinear activation.

\subsection{Relation to existing models}

Our contribution is a nonlinear extension of RPCA for anomaly detection.
As noted above, the key advantages over RPCA are the ability to capture nonlinear structure in the data, as well as the ability to detect anomalies in an inductive setting.
The price we have to pay is the lack of convexity of the objective function, unlike RPCA;
nonetheless, we shall demonstrate that the model can be effectively trained using the procedure described in Section \ref{sec:training}.

Some works have employed deep networks for anomaly detection~\cite{Williams:2002,Zhai:2016},
but without explicitly accounting for gross anomalies.
For example, the recent work of \cite{Zhai:2016} employed an autoencoder-inspired objective to train a probabilistic neural network, with extensions to structured data;
the use of an RPCA-style noise matrix $\N$ may be useful to explore in conjunction with such methods.

Our method is also distinct to denoising autoencoders (DNA), wherein noise is explicitly added to instances~\cite{vincent2010stacked}, whereas we \emph{infer} the noise automatically.
The approaches have slightly different goals: DNAs aim to extract good features from the data, while our aim is to identify anomalies.

Finally, while nonlinear extensions of PCA-style matrix factorisation (including via autoencoders) have been explored in contexts such as collaborative filtering \cite{Lawrence:2009,Sedhain:2015}, we are unaware of prior usage for anomaly detection.

\section{Experimental setup}
\label{sec:experiment-setup}

In this section we show the empirical effectiveness of Robust Convolutional Autoencoder over the state-of-the-art methods on real-world data.
Our primary focus will be on non-trivial image datasets, although our method is applicable in any context where autoencoders are useful e.g.\ speech.

\subsection{Methods compared}
We compare our proposed Robust Convolutional Autoencoder (RCAE)
with the following  state-of-the art methods for anomaly detection:
\let\labelitemi\labelitemii
\begin{itemize}
	\item \textbf{Truncated SVD}, which for zero-mean features is equivalent to PCA. 
	
	\item \textbf{Robust PCA (RPCA)}~\cite{candes2010robust}, as per Equation \ref{eqn:robust-pca}.	 

	\item \textbf{Robust kernel PCA (RKPCA)}~\cite{Nguyen:2009}, as per Equation \ref{eqn:rkpca}.
   	
   	\item \textbf{Autoencoder (AE)}~\cite{bengio2009learning}, as per Equation \ref{eqn:autoencoder}.

   	\item \textbf{Convolutional Autoencoder (CAE)}, a convolutional autoencoder without any accounting for gross anomalies i.e. Equation \ref{eqn:robust-cae} where $\lambda = +\infty$.

   	\item \textbf{Robust Convolutional Autoencoder (RCAE)}, our proposed model as per Equation \ref{eqn:robust-cae}.
\end{itemize}
We used TensorFlow \cite{abadi2016tensorflow} for the implementation of AE, CAE and RCAE\footnote{\url{https://github.com/raghavchalapathy/rcae}}.
For RPCA and RKPCA,
we used
publicly available implementations\footnote{\url{http://perception.csl.illinois.edu/matrix-rank/sample_code.html}}\footnote{\url{http://www3.cs.stonybrook.edu/~minhhoai/downloads.html}}.

\subsection{Datasets}
We compare all methods on three real-world datasets:
\begin{itemize}
	\item {\tt restaurant}, comprising video background modelling and activity detection consisting of snapshots of restaurant activities~\cite{xiong2011direct}.
	\item {\tt usps}, comprising the USPS handwritten digits~\cite{hull1994database}.
	\item {\tt cifar-10} consisting of 60000 $32\times32$ colour images in 10 classes, with 6000 images per class~\cite{krizhevsky2009learning}.
\end{itemize}
For each dataset, we perform further processing to create a well-posed anomaly detection task, as described in the next section.

\subsection{Evaluation methodology}

As anomaly detection is an unsupervised learning problem, model evaluation is challenging.
For the {\tt restaurant} dataset, there are no ground truth anomalies, and so we perform a qualitative analysis by visually comparing the anomalies flagged by various methods, as done in the original robust PCA paper~\cite{candes2010robust}.

For the other two datasets, 
we follow a standard protocol (see e.g.~\cite{xiong2011direct}) wherein anomalies are explicitly identified in the training set.
We can then evaluate the predictive performance of each method as measured against the ground truth anomaly labels,
using three standard metrics:
\begin{itemize}
	\item the area under the precision-recall curve (AUPRC)

	\item the area under the ROC curve (AUROC)

	\item the precision at 10 (P@10).
\end{itemize}
AUPRC and AUROC measure ranking performance, with the former being preferred under class imbalance \cite{Davis:2006}.
P@10 measures classification performance, being the fraction of the top 10 scored instances which are actually anomalous.

For $\tt{CIFAR-10}$,
the labelled dataset is created by combining 5000 images of dogs and 50 images of cats;
a good anomaly detection method should thus flag the cats to be anomalous.
Similarly,
for $\tt{usps}$,
the dataset is created by 
a mixture of 220 images of \lq1\rq s, and 11 images of \lq7\rq as in~\cite{xu2010robust}.
Details of the datasets are summarised in Table \ref{tbl:datasets}.

\begin{table}[!t]
	\centering
	\renewcommand{\arraystretch}{1.25}
	\setlength{\tabcolsep}{6pt}
	\begin{tabular}{@{}llll@{}}
		\toprule
		\toprule
		Dataset & \# instances & \# anomalies & \# features \\
		\toprule
		{\tt restaurant} & 200 & Unknown (foreground) & 19200 \\		
		{\tt usps} 		 & 231 & 11 (\lq7\rq) & 256 \\		
		{\tt cifar-10} 	 & 5000 & 50 (cats) & 1024 \\		
		\bottomrule
	\end{tabular}
	\caption{Summary of datasets used in experiments.}
	\label{tbl:datasets}
\end{table}

Additionally,
we also test the ability of our model to perform denoising of images,
as well as detecting inductive anomalies.

\subsection{Network parameters}

Although we have observed that deeper RCAE networks tend to achieve better image reconstruction performance, there exist four fold options related to network parameters to be chosen:
(a) number of convolutional filters, (b) filter size, (c) strides of convolution operation and (d) activation applied.
We tuned via grid search additional hyper-parameters, including the number of hidden-layer nodes $H \in \{3, 64, 128\}$, and regularisation $\lambda$ within range ${[0, 100]}$.
The learning, drop-out rates and regularization parameter $\mu$ were sampled from a uniform distribution in the range $[0.05, 0.1]$.
The embedding and initial weight matrices were all sampled from the uniform distribution within range $[-1, 1]$.

\section{Experimental results}
\label{sec:experiment-results}

In this section, we present experiments for three scenarios:
(a) non-inductive anomaly detection,
(b) inductive anomaly detection, and
(c) image denoising.

\subsection{Non-inductive anomaly detection results}

We present results on the three datasets described in Section \ref{sec:experiment-setup}.

\subsubsection{{(1) {\tt restaurant} dataset}}
We work with the {\tt restaurant} video activity detection dataset~\cite{xiong2011direct},
and consider the problem of inferring the background of videos via removal of (anomalous) foreground pixels.
Estimating the background in videos is important for tasks such as anomalous activity detection.
It is however difficult because of the variability of the background (e.g. due to lighting conditions) and the presence of foreground objects such as moving objects and people.

For this experiment, we only compare the RPCA and RCAE methods, owing to a lack of ground truth labels.

\textbf{Parameter settings}.
For RPCA, rank $K$ = 64 is used.

Per the success of the Batch Normalization architecture~\cite{ioffe2015batch} and Exponential Linear Units~\cite{clevert2015fast}, we have found that convolutional+batch-normalization+elu layers provide a better representation of convolutional filters.
Hence, in this experiment the RCAE adopts four layers of (conv-batch-normalization-elu) in the encoder part and four layers of  (conv-batch-normalization-elu) in the decoder portion of the network.
RCAE network parameters such as (number of filter, filter size, strides) are chosen to be (16,3,1) for first and second layers and (32,3,1) for third and fourth layers of both encoder and decoder layers.

\begin{figure}[!t]
	\centering
	\subfigure[RCAE.]{\includegraphics[scale=0.325]{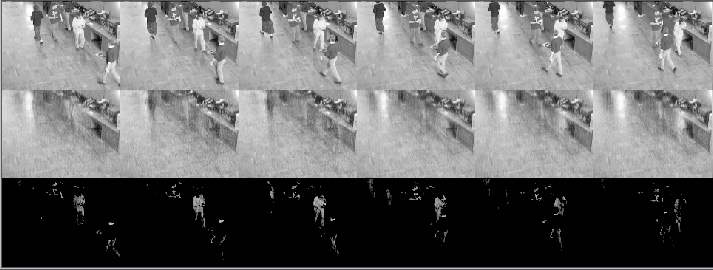}}
	\subfigure[RPCA.]{\includegraphics[scale=0.325]{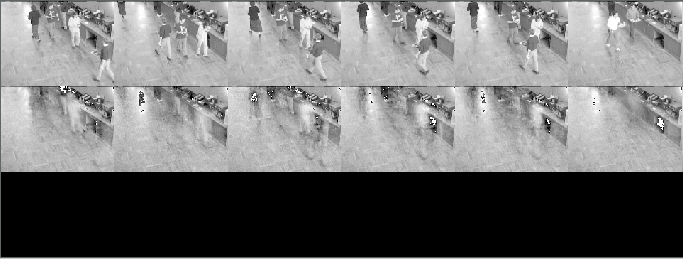}}
	\caption{Top anomalous images containing original image (people walking in the lobby) decomposed into background (lobby) and foreground (people), {\tt restaurant} dataset.}
	\label{fig:results-restaurant}
\end{figure}
\textbf{Results}.
While there are no ground truth anomalies in this dataset, a qualitative analysis reveals RCAE to outperforms its counterparts in capturing the foreground objects.
Figure~\ref{fig:results-restaurant} compares the top 6 most anomalous images for RCAE and RPCA.
We see that the most anomalous images contain high foregound activity (which are recognised as anomalous).
Visually, we see that the background reconstruction produced by RPCA contains a few blemishes in some cases, while for RCAE the backgrounds are smooth.

\subsubsection{{(2) {\tt usps} dataset}}
From the {\tt usps} handwritten digit dataset,
we create a dataset 
with a mixture of 220 images of \lq1\rq s, and 11 images of \lq7\rq, as in~\cite{xu2010robust}.
Intuitively, the latter images are treated as being anomalous, as the corresponding images have different characteristics to the majority of the training data. Each image is flattened as a row vector, yielding a 231 $\times$ 256 training matrix.

\textbf{Parameter settings}.
For SVD and RPCA methods, rank $K = 64$ is used.
For AE, the inputs are flattened images as a column vector of size 256,
and the hidden layer is a column vector of size  64 (matching the rank $K$).

For DRMF, we follow the settings of~\cite{xu2010robust}.
For RKPCA, we used a Gaussian kernel with bandwidth $0.01$, a cost parameter $C = 1$, and requested $60\%$ of the KPCA spectrum (which roughly selects 64 principal components).

For RCAE, we set two layers of convolution layers with the filter number to be 32, filter size to be 3$\times$3, with number of strides as $1$ and  rectified linear unit (ReLU) as activation with max-pooling layer of dimension 2$\times$2.

\begin{figure}[!t]
	\centering
	\subfigure[RCAE.]{\includegraphics[scale=0.9]{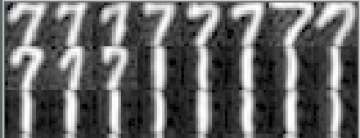}}
	\quad
	\subfigure[RPCA.]{\includegraphics[scale=0.29]{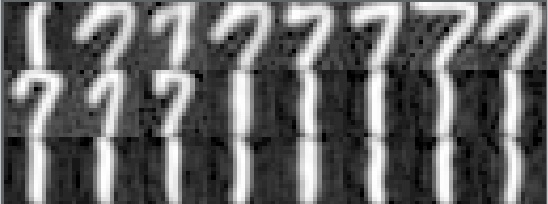}}
	\caption{Top anomalous images, {\tt usps} dataset.}
	\label{fig:usps-anomalies}
\end{figure}

\textbf{Results}.
From Table~\ref{tbl:anomaly-results-summary}, we see that it is a near certainty for all \lq7\rq\, are accurately identified as outliers.
Figure~\ref{fig:usps-anomalies} shows the top anomalous images for RCAE, where indeed the \lq7\rq's are correctly placed at the top of the list.
By contrast, for RPCA there are also some \lq1\rq's placed at the top.

\begin{table*}[!t]	
	\centering
	\renewcommand{\arraystretch}{1.25}
	\resizebox{0.99\linewidth}{!}{
	\subfigure[{\tt usps}]{
		\begin{tabular}{lccc}
			\toprule
			\toprule
			\textbf{Methods} & \textbf{AUPRC} & \textbf{AUROC} & \textbf{P@10} \\
			\midrule
			RCAE  & \cellcolor{gray!25}{0.9614 $\pm$ 0.0025}&\cellcolor{gray!25}{0.9988$\pm$ 0.0243}&\cellcolor{gray!25}{0.9108 $\pm$ 0.0113} \\
			\midrule
			CAE & 0.7003 $\pm$ 0.0105 & 0.9712 $\pm$ 0.0002 & 0.8730 $\pm$ 0.0023\\
			AE  & 0.8533 $\pm$ 0.0023 & 0.9927 $\pm$ 0.0022 & 0.8108 $\pm$ 0.0003 \\		
			\midrule
			RKPCA & 0.5340 $\pm$ 0.0262 & 0.9717 $\pm$ 0.0024 & 0.5250 $\pm$ 0.0307 \\
			DRMF  & 0.7737 $\pm$ 0.0351 & 0.9928 $\pm$ 0.0027 & 0.7150 $\pm$ 0.0342 \\
			RPCA  & 0.7893 $\pm$ 0.0195 & 0.9942 $\pm$ 0.0012 & 0.7250 $\pm$ 0.0323\\
			SVD   & 0.6091 $\pm$ 0.1263 & 0.9800 $\pm$ 0.0105 & 0.5600 $\pm$ 0.0249 \\
			\bottomrule	
	\end{tabular}}%
	\quad
	\subfigure[{\tt cifar-10}]{
		\begin{tabular}{ccc}
			\toprule
			\toprule
			\textbf{AUPRC} & \textbf{AUROC} & \textbf{P@10} \\
			\midrule
			\cellcolor{gray!25}{0.9934 $\pm$ 0.0003}&\cellcolor{gray!25}{0.6255 $\pm$ 0.0055} &\cellcolor{gray!25}{0.8716 $\pm$ 0.0005} \\
			\midrule
			0.9011 $\pm$ 0.0000 & 0.6191 $\pm$ 0.0000 & 0.0000 $\pm$ 0.0000 \\
			0.9341 $\pm$ 0.0029 & 0.5260 $\pm$ 0.0003 & 0.2000 $\pm$ 0.0003 \\
			\midrule			
			0.0557 $\pm$ 0.0037 & 0.5026 $\pm$ 0.0123 & 0.0550 $\pm$ 0.0185 \\
			0.0034 $\pm$ 0.0000 & 0.4847 $\pm$ 0.0000 & 0.0000 $\pm$ 0.0000 \\
			0.0036 $\pm$ 0.0000 & 0.5211 $\pm$ 0.0000 & 0.0000 $\pm$ 0.0000 \\
			0.0024 $\pm$ 0.0000 & 0.5299 $\pm$ 0.0000 & 0.0000 $\pm$ 0.0000 \\
			\bottomrule	
	\end{tabular}}%
	}	

	\caption{Comparison between the baseline (bottom four rows) and state-of-the-art systems (top three rows). Results are the mean and standard error of performance metrics over 20 random training set draws. Highlighted cells indicate best performer.}
	\label{tbl:anomaly-results-summary}
\end{table*}

\subsubsection{{(3) {\tt cifar-10} dataset}}
\begin{figure}[!t]
	\centering
	\subfigure[RCAE.]{\includegraphics[scale=0.95]{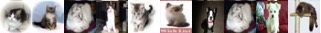}}
	\subfigure[RPCA.]{\includegraphics[scale=0.95]{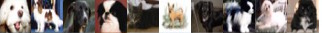}}
	\caption{Top anomalous images, 
	{\tt cifar-10} dataset.}
	\label{fig:results-cifar}
\end{figure}

We create a dataset with anomalies
by combining 5000 random images of dogs and 50 images of cats, as illustrated in Figure~\ref{fig:results-cifar}.
In this scenario the cats are anomalies, and the goal is to detect all the cats in an unsupervised manner.

\textbf{Parameter settings}.
For SVD and RPCA methods, rank $K = 64$ is used.
We trained a three-hidden-layer autoencoder (AE) (1024-256-64-256-1024 neurons).
The middle hidden layer size is set to be same as rank $K = 64$, 
and the model is trained using Adam~\cite{kingma2014adam}.
The decoding layer uses sigmoid function in order to capture the nonlinearity characteristics from  latent representations produced by the hidden layer.
Finally, we obtain the feature vector for each image by obtaining the latent representation from the hidden layer. 

For RKPCA, we used a Gaussian kernel with bandwidth $5 \cdot 10^{-8}$, a cost parameter $C = 0.1$, and requested $55\%$ of the KPCA spectrum (which roughly selects 64 principal components).
The RKPCA runtime was prohibitive on the full sample (see Sec \ref{sec:runtime}), so we resorted to a subsample of 1000 dogs and 50 cats.

The RCAE architecture in this experiment is same as for {\tt restaurant}, containing 
four layers of  (conv-batch-normalization-elu) in the encoder part
and four layers of  (conv-batch-normalization-elu) in the decoder
portion of the network. RCAE network parameters such as (number of filter, filter size, strides) are chosen to be (16,3,1)  for first and second layers and (32,3,1) for third and fourth layers of both encoder and decoder.

\textbf{Results}.
From Table \ref{tbl:anomaly-results-summary},
RCAE clearly outperforms all existing state-of-the art methods in anomaly detection.
Note that basic CAE, with no robustness (effectively $\lambda = \infty$), is also outperformed by our method, indicating that it is crucial to explicitly handle anomalies with the $\N$ term.

Figure~\ref{fig:results-cifar} illustrates the most anomalous images for our RCAE method, compared to RPCA.
Owing to the latter involving learning a linear subspace, the model is unable to effectively distinguish cats from dogs;
by contrast, RCAE can effectively determine the manifold characterising most dogs, and identifies cats to be anomalous with respect to this.

\subsection{Inductive anomaly detection results}

We conduct an experiment to assess the detection of \emph{inductive} anomalies.
Recall that this is a capability of our RCAE model, but not e.g. RPCA.
We consider the following setup:
we train our model on 5000 dog images, and then evaluate it on a test set comprising 500 dogs and 50 cat images.
As before, we wish all methods to accurately determine the cats to be anomalies.

\begin{figure}[!t]
	\centering
	\subfigure[RCAE.]{\includegraphics[scale=0.31]{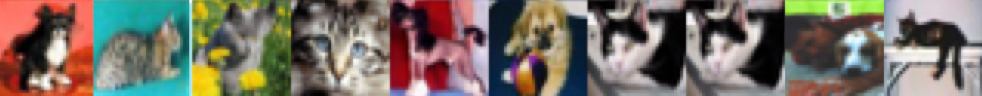}}
	\subfigure[CAE.]{\includegraphics[scale=0.95]{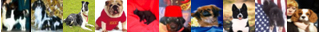}}
	\caption{Top inductive anomalous images, {\tt cifar-10} dataset.}
	\label{fig:results-cifar-inductive}
\end{figure}

Table~\ref{tbl:inductive-anomaly-results} summarises the detection performance for all the methods on this inductive task.
The lower values compared to Table~\ref{tbl:anomaly-results-summary} are indicative that the problem here is more challenging than anomaly detection on a single dataset;
nonetheless, we see that our RCAE method manages to convincingly outperform both the SVD and AE baselines.
This is confirmed qualitatively in Figure~\ref{fig:results-cifar-inductive}, where we see that RCAE correctly identifies many cats in the test set as anomalous, while the basic
CAE
method suffers.

\begin{table}[!t]
	\centering
	\renewcommand{\arraystretch}{1.25}	
	\caption{Inductive anomaly detection results on {\tt cifar-10}. Note that RPCA and DRMF are inapplicable here. Highlighted cells indicate best performer.}	
	\resizebox{0.99\linewidth}{!}{
	\begin{tabular}{@{}lccccc@{}}
		\toprule
		\toprule
		& \textbf{SVD} & \textbf{RKPCA} & \textbf{AE} & \textbf{CAE} & \textbf{RCAE} \\
		\toprule
		\bf{AUPRC} & 0.1752 $\pm$ 0.0051 & 0.1006 $\pm$ 0.0045 & 0.6200 $\pm$ 0.0005 & 0.6423 $\pm$ 0.0005 & \cellcolor{gray!25}{0.6908 $\pm$ 0.0001} \\	
		\bf{AUROC} & 0.4997 $\pm$ 0.0066 & 0.4988 $\pm$ 0.0125 & 0.5007 $\pm$ 0.0010 & 0.4708 $\pm$ 0.0003 & \cellcolor{gray!25}{0.5576 $\pm$ 0.0005} \\
		\bf{P@10}  & 0.2150 $\pm$ 0.0310 & 0.0900 $\pm$ 0.0228 & 0.1086 $\pm$ 0.0001 & 0.2908 $\pm$ 0.0001 & \cellcolor{gray!25}{0.5986 $\pm$ 0.0001} \\
		\bottomrule
	\end{tabular}	
	}	
	\label{tbl:inductive-anomaly-results}
\end{table}

\subsection{Image denoising results}

Finally, we test the ability of the model to de-noise images, which is a form of anomaly detection on individual pixels (or more generally, features).
In this experiment, we train all models on a set of 5000 images of dogs from {\tt cifar-10}.
For each image, we then add salt-and-pepper noise at a rate of 10\%.
Our goal is to recover the original image as accurately as possible.

\begin{figure}[!t]
	\centering
	\subfigure[RCAE.]{\includegraphics[scale=0.95]{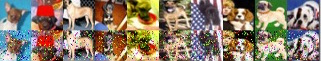}}
	\subfigure[RPCA.]{\includegraphics[scale=0.95]{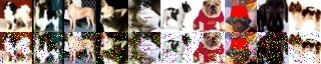}}
	\caption{Top anomalous images in original form (first row), noisy form (second row),
	image denoising task on {\tt cifar-10}.}
	\label{fig:results-cifar-injection}
\end{figure}

\begin{figure}[!t]
	\centering
	\includegraphics[scale=0.5]{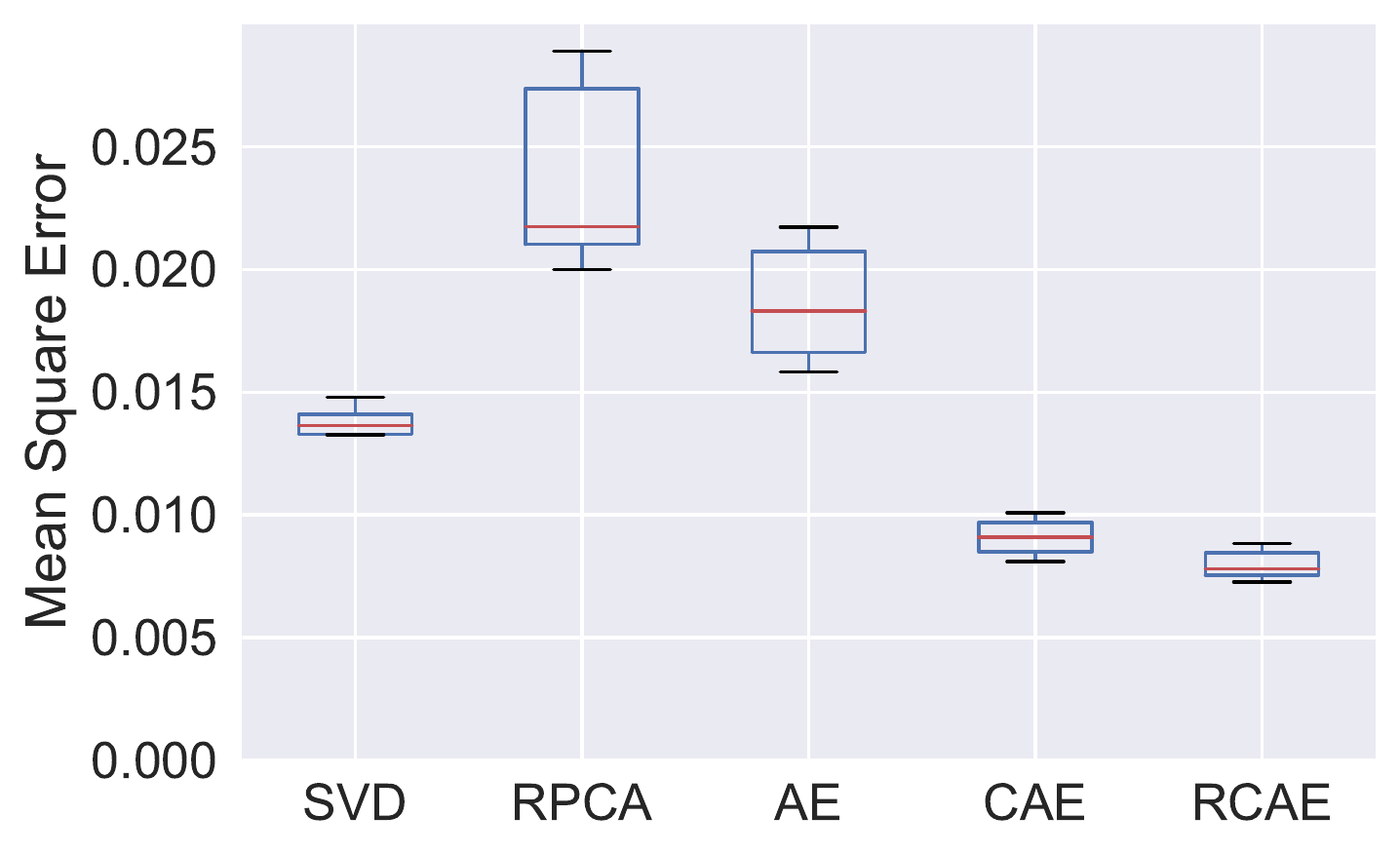}
	\caption{Illustration of the mean square error boxplots obtained for various models on image denoising task, {\tt cifar-10} dataset.
		In this setting, RCAE suppresses the noise and detects the background and foreground images effectively.}
	\label{fig:denoising-results}
\end{figure}

Figure~\ref{fig:results-cifar-injection} illustrates that the most anomalous images in the presence of noise
contain images of the variations of dog class images (e.g. containing person's face).
Further, Figure \ref{fig:denoising-results} illustrates for various methods the mean square error between the reconstructed and original images.
RCAE effectively suppresses the noise as evident from the low error.
The improvement over raw CAE is modest, but suggests that there is benefit to explicitly accounting for noise.

\subsection{Comparison of training times}
\label{sec:runtime}

We remark finally that our RCAE method is comparable in training efficiency to existing methods.
For example, on the small-scale {\tt restaurant} dataset, it takes 1 minute to train RPCA,
and 8.5 minutes to train RKPCA,
compared with 10 minutes for our RCAE method.
The ability to leverage recent advances in deep learning as part of our optimisation (e.g.\ training models on a GPU) is we believe a salient feature of our approach.

We note that while the RKPCA method is fast to train on smaller datasets, on larger datasets it suffers from the $O(n^2)$ complexity of kernel methods;
for example, it takes over an hour to train on the {\tt cifar-10} dataset.
It is plausible that one could leverage recent advances in fast approximations of kernel methods~\cite{Lopez-Paz:2014}, and studying these would be of interest in future work.
Note that the issue of using a fixed kernel function would remain, however.

\section{Conclusion}
\label{sec:conclusion}
We have extended the robust PCA model to the nonlinear autoencoder setting.
To the best of our knowledge, ours is the first approach which is \emph{robust}, \emph{nonlinear} and \emph{inductive}.
The robustness ensures that the model is not over-sensitive
to anomalies;
the nonlinearity helps discover potentially more
subtle anomalies;
and being inductive makes it possible
to deploy our model in a live setting.

While autoencoders are a powerful mechansim for data representation
they suffer from their ``black-box'' nature. There is a growing
body of research on outlier description, i.e., explain the reason
why a data point is anomalous. A direction of future reason is 
to extend deep autoencoders for outlier \emph{description}.

\bibliographystyle{splncs03}
\bibliography{MyBibFile,outlier}

\end{document}